\pdfoutput=1
\documentclass[11pt]{article}

\usepackage[]{EACL2023}
\usepackage{nameref}
\usepackage{times}
\usepackage{latexsym}

\usepackage[T1]{fontenc}
\usepackage[utf8]{inputenc}

\usepackage{microtype}

\usepackage{inconsolata}
\usepackage{booktabs}
\usepackage{graphicx}

\usepackage{array}
\usepackage{enumitem}
\usepackage{amsmath,amssymb}
\usepackage{colortbl}
\usepackage{multirow}
\title{What happens \emph{before} and \emph{after}:\\Multi-Event Commonsense in Event Coreference Resolution}

\author{Sahithya Ravi$^{1,2}$~~~Chris Tanner$^{3,4}$~~Raymond Ng$^{1}$ ~~Vered Shwartz$^{1,2}$ \\
$^1$ University of British Columbia\\$^2$ Vector Institute for AI\\$^3$ Massachusetts Institute of Technology\\$^4$ Kensho Technologies\\
{\tt\small \{sahiravi, rng, vshwartz\}@cs.ubc.ca, cwt@mit.edu }}
\newcommand{\eg}{{\em e.g.}}

\newcommand{\quotes}[1]{``#1''}
\newcolumntype{P}[1]{>{\raggedright\arraybackslash}p{#1}}

\begin{document}
\maketitle
\begin{abstract}
Event coreference models cluster event mentions pertaining to the same real-world event. Recent models rely on contextualized representations to recognize coreference among lexically or contextually similar mentions. However, models typically fail to leverage commonsense inferences, which is particularly limiting for resolving lexically-divergent mentions. We propose a model that extends event mentions with temporal commonsense inferences. Given a complex sentence with multiple events, e.g., ``The man killed his wife and got arrested'', with the target event ``arrested'', our model generates plausible events that happen before the target event – such as ``the police arrived'', and after it, such as ``he was sentenced''. We show that incorporating such inferences into an existing event coreference model improves its performance, and we analyze the coreferences in which such temporal knowledge is required.

\end{abstract}

\section{Introduction}
\label{sec:intro}
% First paragraph: event coreference motivation
The goal of cross-document event coreference resolution is to determine if various event mentions (e.g. \textit{shot}, \textit{gunshot}), across one or more documents, refer to the same event. Existing systems represent each mention within its context using a language model \cite{cattan-etal-2021-cross-document,allaway-etal-2021-sequential}, and train a scorer to predict if two mentions corefer, based on their lexical and contextual similarity. 

\begin{figure*}
    \centering
    \includegraphics[width=\textwidth]{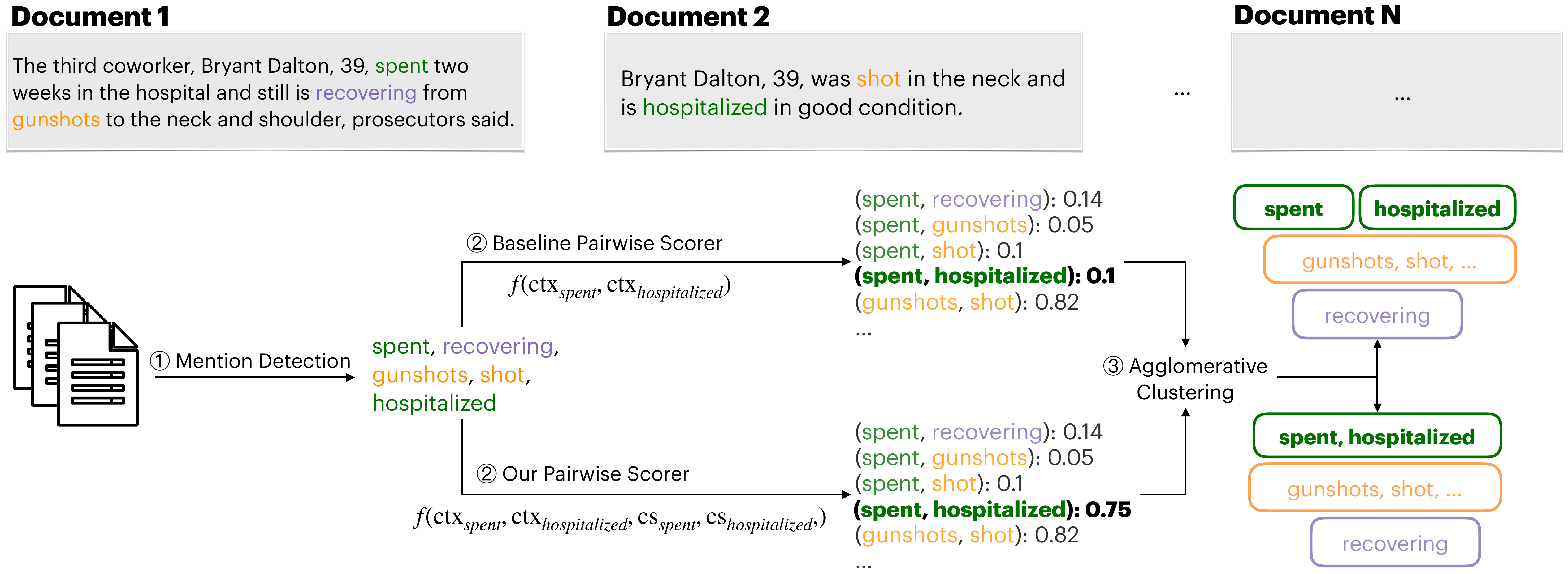}
    \caption{
The architecture of our event coreference model, exemplified on a set of documents. We enhance the pairwise scorer, originally based on contextual similarity (top), with temporal commonsense with temporal commonsense inference embeddings – indicated with \emph{cs} (bottom). Such inferences help identify lexically divergent co-referring pairs, such as \textit{spent} [time at a hospital] and \textit{hospitalized}.}
    \label{fig:architecture}
\end{figure*}

While many coreferring mention pairs in event coreference datasets such as ECB+ \cite{cybulska-vossen-2014-using} are lexically and contextually similar, or even share the same lemma \cite{wolfe-etal-2015-predicate}, the difficulty arises for dissimilar coreferring mentions. For example, in Figure~\ref{fig:architecture}, \textit{spent} and \textit{hospitalized} are coreferring. These mentions are not lexically similar, and are not often used in similar contexts. In this paper, we improve the ability of existing cross-document event coreference systems to resolve such challenging coreferring mentions, by providing additional context in the form of commonsense knowledge. We focus on two temporal commonsense relations --- \emph{before} and \emph{after} --- pertaining to typical events that happen before and after the target event. For instance, in Figure~\ref{fig:architecture}, we may infer that before Dalton was \textit{shot}, a shooter loaded their gun. Similarly, we may infer that Dalton was hurt prior to his \textit{hospitalization} and got discharged afterward.
% Fourth paragraph: the multi-event model
Our first contribution is the development of a commonsense reasoning engine that can reason about these two temporal relations. Existing commonsense models \cite{paracomet, Hwang2021COMETATOMIC2O} may generate such inferences for simple sentences with a single event, such as ``Bryant Dalton was shot'', but they do not support complex sentences with multiple events of interest (e.g. \textit{shot}, \textit{hospitalized}). Further, they may conflate the inferences for different events. We develop a multi-event commonsense model that considers the entire context and is capable of generating separate inferences for each target event in complex sentences.

As an additional contribution, we incorporate the inferences into the pairwise mention scorer of a cross-document event coreference system \cite{cattan-etal-2021-cross-document}. We produce \emph{before} and \emph{after} inferences for each event mention. We then embed the inferences, either by attending each mention to its own inferences (intra-span) or to the other mention's inferences (inter-span). 

The results confirm that commonsense inferences are useful for event coreference. Each of our model variants improves upon the baseline performance, with the intra-span version performing the best. We analyze the successful predictions and interpret how the commonsense inferences help resolve difficult mention pairs.
%and perform a detailed error analysis. 
In the future, we plan to extend our multi-event inference engine to additional commonsense knowledge types and apply it to other discourse tasks, such as summarization, dialogue, and story comprehension.\footnote{The code is available \href{https://github.com/sahithyaravi1493/coref_csk}{here.}}

\section{Background}
\label{sec:related_work}
In this work, we improve the performance of a system for cross-document coreference resolution by incorporating temporal commonsense inferences pertaining to the events. We first provide background on event coreference resolution (\S\ref{sec:bg:coref}). We then describe related work concerning event-centered commonsense (\S\ref{sec:bg:event_commonsense}), along with approaches for using language models for data augmentation (\S\ref{sec:bg:syntheticdata}).

\subsection{Event Coreference Resolution}
\label{sec:bg:coref}
Event coreference resolution aims to cluster event mentions that refer to the same underlining real-world occurrence.
Our work focuses on cross-document coreference resolution (CD), which aims to resolve mentions across an entire corpus of documents. In contrast, the problem of within-document coreference resolution (WD) only resolves mentions on a per-document basis. Event coreference is often performed jointly with \textit{entity} coreference resolution, which concerns resolving mentions of people, locations, and organizations. 
\paragraph{Datasets.}
In this paper, we use the ECB+ dataset proposed by \citet{cybulska-vossen-2014-using} and widely accepted as the standard benchmark for coreference resolution (CD). ECB+ contains 86 sub-topics, each of which concerns a specific news event.
To introduce complexity and difficulty, each sub-topic is highly similar to -- yet distinctly different from -- exactly one other sub-topic. ECB+ includes both entity and event mentions; however, this work focuses solely on events. Table~\ref{tab:dataset} shows event statistics from ECB+ corpus.
% For example, one sub-topic concerns \quotes{Ellen hosting the Oscars}, whereas another sub-topic concerns \quotes{Jackman hosting the Academy awards}.
\begin{table}[t]
\centering
\small
\begin{tabular}{@{}lllll@{}}
\toprule
\textbf{}      & \textbf{Train} & \textbf{Dev} & \textbf{Test} & \textbf{} \\ \midrule
Topics (subtopic-pairs)        & 25             & 8            & 10            &           \\
Event mentions & 3808           & 1245         & 1780          &           \\
Event clusters & 1527           & 409          & 805           &           \\ \bottomrule
\end{tabular}
\caption{Statistics on the standard train/dev/test split of event coreferences in ECB+ dataset \cite{cybulska-vossen-2014-using}}
\label{tab:dataset}
\end{table}
\paragraph{Models.} Recent approaches to CD event coreference often follow the architecture described in Figure~\ref{fig:architecture}.  First, candidate event mentions are extracted from documents. Second, a pairwise scorer is trained to classify every pair of mentions as being coreferent or not. Finally, these scores are used to form distinct clusters of event mentions, typically using agglomerative clustering. Amongst these components, coreference models tend to mostly vary in their scoring approach (i.e., second component).\mbox{} \\
Early approaches relied on lexical and syntactic features  \cite{yang-etal-2015-hierarchical,choubey-huang-2017-event}, or used semantic roles to encode the relationships between entities and events. Recently, \newcite{meged-etal-2020-paraphrasing} improved performance by leveraging a resource of predicate paraphrases. Finally, \newcite{lai-etal-2021-context} incorporated entities, relations, and events extracted from a state-of-the-art information extraction system. Generally, current state-of-the-art models often rely on pre-trained language models to compute a contextualized representation for each candidate mention, which serve as input to the pairwise scorer \cite[e.g.][]{contextual2,zeng-etal-2020-event, cattan-etal-2021-cross-document,allaway-etal-2021-sequential}. 

Our model is an enhancement of the model proposed by \newcite{cattan-etal-2021-cross-document}. It targets both entity and event coreference resolution, and in an end-to-end fashion it performs mention extraction, pairwise scoring, and clustering (Figure~\ref{fig:architecture}). Mentions are represented by contextualized embeddings from RoBERTa \cite{roberta-large}. We chose to base our model on \newcite{cattan-etal-2021-cross-document} for two reasons. First, it is a simple model following the standard approach presented in Figure~\ref{fig:architecture}. Later approaches rely on hierarchical representations \cite{yadav-etal-2021-event} or discourse coherence theory \cite{held-etal-2021-focus}. Second, it is based on RoBERTa and is more efficient and less memory consuming than the succeeding CDLM model \cite{caciularu-etal-2021-cdlm-cross} that is based on the much larger Longformer model \cite{Beltagy2020Longformer}. 
\mbox{} 

More recently, \newcite{yadav-etal-2021-event} built on \newcite{cattan-etal-2021-cross-document} by proposing a hierarchical approach to representing uncertainty of clustering event and entity mentions. The state-of-the-art models for cross document coreference are \newcite{caciularu-etal-2021-cdlm-cross}, which models cross-text relationships by using larger context windows, and \newcite{held-etal-2021-focus}, which applies discourse coherence theory to coreference.

\subsection{Event-Centric Commonsense}
\label{sec:bg:event_commonsense}
Commonsense reasoning helps humans bridge the gap between utterance and intended meaning. Reasoning about events has long been of interest to AI research. \newcite{schank1975scripts} introduced ``scripts'' as a prototypical series of events, e.g. going to a restaurant is composed of ordering food, eating, and paying, and the participants: customer, waiter, and cook. Various methods have been proposed to learn such scripts from text \cite[e.g.][]{chambers-jurafsky-2008-unsupervised,pichotta-mooney-2014-statistical,rudinger-etal-2015-script}. 

The ATOMIC knowledge base \cite{Hwang2021COMETATOMIC2O, sap2019atomic} consists of 1.1M crowdsourced event-relation-event triplets pertaining to the causes, effects, and mental states of the event participants. To generate contextually-relevant ATOMIC-style inferences, 
%dynamically for new inputs, 
\newcite{bosselut-etal-2019-comet} developed COMET, a pre-trained language model fine-tuned on ATOMIC. COMET has shown promising results on tasks such as therapy chatbots \cite{kearns2020wizard}, persona-grounded dialogue \cite{majumder-etal-2020-like}, figurative language interpretation and generation \cite{chakrabarty-etal-2020-r,chakrabarty-etal-2022-rocket}, and temporal ordering of sentences \cite{ghosal-etal-2021-stack}. 

Several variants of COMET have been subsequently released. ParaCOMET \cite{paracomet} adapts COMET to generate sentence-level inferences within the context of an entire paragraph. VisualCOMET \cite{visualcomet} generates ATOMIC-style inferences for images. Finally, the updated version of COMET  \cite{Hwang2021COMETATOMIC2O} extends the relation inventory and crowdsources more inferences. The additional inferences include the two temporal relations that are the most relevant to our work, ``happens before'' and ``happens after''.

\subsection{LM-generated Data Augmentation}
\label{sec:bg:syntheticdata}

The success of using large pre-trained LMs in a few-shot setup for generation tasks has led to an increased interest in using such models to generate data for downstream tasks. Recent work augmented datasets by fine-tuning a pre-trained LM on real data, then generated new, silver-labelled instances \cite{AnabyTavor2020DoNH,Papanikolaou2020DAREDA,kumar-etal-2020-data}. Similarly, the few-shot capabilities of GPT-3 \citep{gpt3} were leveraged to generate free-text explanations \cite{wiegreffe-etal-2022-reframing}, semantically-related sentence pairs \cite{Schick2021GeneratingDW}, atomic event commonsense triples \cite{West2022SymbolicKD}, and labels for various generation and understanding tasks \cite{wang-etal-2021-want-reduce}. In this work, we fine-tune GPT-3 with minimal human supervision to generate additional contextual data pertaining to events.

%With the success of large pre-trained language models such on few-shot generation there has been an increasing interest in using such models to generate synthetic data for downstream tasks. Some recent research look into approaches to increase the size of datasets by fine-tuning pre-trained language models on human annotations \citep{AnabyTavor2020DoNH,Papanikolaou2020DAREDA,kumar-etal-2020-data}. Similarly, the few-shot capabilities of GPT models\citep{gpt3} has recently been utilized in \citet{wiegreffe-etal-2022-reframing} generating free-text explanations, \cite{Schick2021GeneratingDW} for generating semantically related sentence pairs \citet{West2022SymbolicKD} for generating single-event commonsense triples and \cite{wang-etal-2021-want-reduce} for generating labels for natural language generation and understanding tasks. In this work, we use GPT-3 \cite{gpt3} we fine-tune GPT-3 with minimal human supervision to generate additional contextual data related to events.

\section{Method}
\label{sec:method}
\begin{figure*}[t]
    \centering
    \small
    \begin{tabular}{|P{0.8\textwidth}|}
    \hline
    \textbf{Instructions:} Read the context sentence and write at least \textbf{two} inferences for question 1 and \textbf{two} inferences for question 2. As shown in the examples, each inference is expected to be a short sentence between \textbf{5-10 words}.\\ \hline
    \textbf{Context:} A publicist says Tara Reid has \textbf{checked herself} into rehab.\\
    \textbf{Question 1:} What typically happens \textcolor{orange}{before} the event \textcolor{blue!70}{checked herself}? \\
    \textbf{Question 2:} What typically happens \textcolor{teal}{after} the event \textcolor{blue!70}{checked herself}? \\ \hline
    \end{tabular}
    \caption{An example task on Amazon Mechanical Turk.}
    \label{fig:mturk}
\end{figure*}

The architecture of our method is shown in Figure~\ref{fig:architecture}. We use the same clustering method as in \newcite{cattan-etal-2021-cross-document} but revise the pairwise scorer. Our goal is to improve the model's ability to resolve coreferences between mention pairs that are not lexically or contextually similar, but where one mention could be inferred from the other using commonsense knowledge and reasoning. Thus, we develop a commonsense inference engine (Sec~\ref{sec:method:csmodel}) and use it to enhance the pairwise scorer (Sec~\ref{sec:method:incorporation}). 

% We focus on the pairwise scorer of the coreference resolution pipeline and adapt it to incorporate commonsense knowledge from our event commonsense model. In this section, we describe our approach in detail, starting with collecting annotations (Sec \ref{method:crowdsource}), building an event commonsense model (Sec \ref{method:csmodel}), followed by incorporating the inferences from the commonsense model (Sec \ref{method:incorporation}).

\subsection{Multi-Event Commonsense Inferences}
\label{sec:method:csmodel}

We enhance the pairwise scorer with commonsense inferences regarding the events' temporal aspects. Specifically, we focus on plausible events that might have happened before or after the target event. For example, in Figure~\ref{fig:architecture}, after being hospitalized, the victim received treatment. 

We found COMET and its variants to be ineffective for generating inferences for our task. COMET was trained on the ATOMIC knowledge base \cite{sap2019atomic}. As the name implies, events are atomic, i.e., comprise a single verb phrase. Conversely, the existing event coreference datasets are based on news articles, where sentences often contain multiple events. COMET predictions for document 1 (Figure~\ref{fig:architecture}) have no indication which verb they pertain to. Moreover, COMET predicts that what happens after document 1 is murder, which contradicts the fact that the victim survived and was taken to the hospital. ParaCOMET \cite{gabriel-etal-2021-discourse} facilitates generating consistent inferences for multi-sentence paragraphs, but it was trained on the ROCStories dataset \cite{mostafazadeh-etal-2016-corpus}, which is in the fiction domain and in which sentences are also simple. 

% We found COMET and its variants to be ineffective for generating commonsense for the task of event coreference resolution. This is due to the fact that commonsense models like COMET are trained on atomic events (on ATOMIC), whereas most coreference and other discourse tasks involve multi-event sentences. This indicated that we require a more complex model that can draw commonsense inferences on particular events in the sentence.

To that end, we trained a new multi-event commonsense inference engine. Given a sentence with multiple events (such as document 1), and a target event (e.g. \textit{hospitalized}), the goal is to generate what might have happened before and after the target event---in the context of the entire sentence. 

\paragraph{Model.} We base the inference engine on GPT-3 \cite{gpt3}. While GPT-3 is not directly applicable to the task of event coreference \cite{yang-etal-2022-gpt}, it has been shown to contain a wealth of factual and commonsense knowledge as a result of extensive pre-training. Our goal is to use this knowledge to generate event-centric commonsense inferences without requiring extensive training. GPT-3 is especially well-suited for this task, as it has shown remarkable performance in learning from fewer examples in a variety of tasks.

\begin{figure*}
    \centering
    \includegraphics[width=\textwidth]{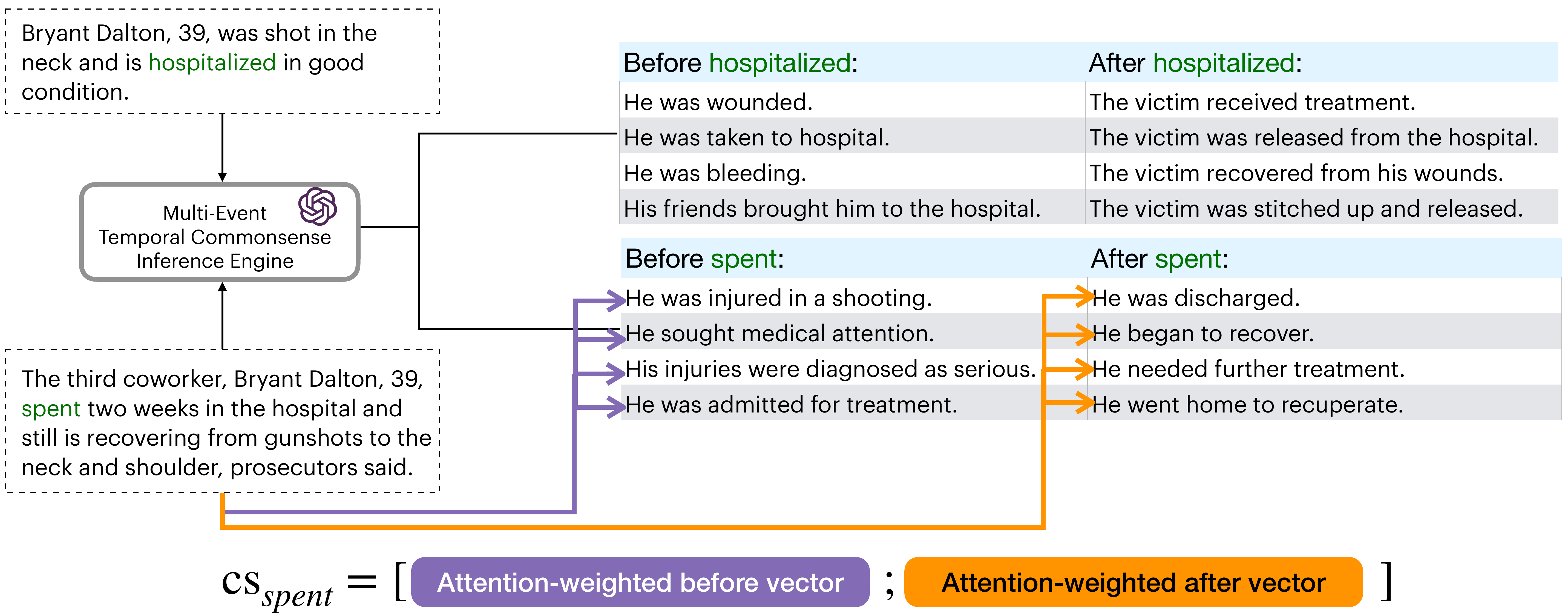}
    \caption{An illustration of the new additions to the pairwise scorer. We input each document into a GPT-3-based multi-event temporal commonsense inference engine, which outputs plausible events that happened before and after the target event (e.g. \textit{spent}). For each temporal relation (i.e., \emph{before} and \emph{after}), we embed the corresponding inferences and compute an attention-weighted vector. We concatenate the \emph{before} and \emph{after} vectors to the mention representations as input to the pairwise scorer.}
    \label{fig:pairwise_scorer}
\end{figure*}

\paragraph{Data.} As the first step in training a multi-event commonsense model, we crowdsourced annotations for 100 events -- using the gold standard event mentions from the ECB+ training set. 
To include a wide range of topics, we selected the first four events from each of the 25 topics in the training set.
%From each of the 25 topics in the training set, we selected the first 4 events, covering a broad range of topics. 

We presented workers with a sentence with one or more events, and asked them to describe what happens immediately before and after the target event. Figure~\ref{fig:mturk} shows an example\footnote{See Appendix~\ref{appendix:he} for the exact template.}. We obtained annotations from three workers for each sentence, and instructed workers to write at least two inferences for each relation. This yielded a total of 600 inferences ($100 \times 3 \times 2 = 600$). We carefully reviewed the data and removed a handful of inferences that were of poor quality (i.e., incomplete or irrelevant sentences, which amounted to roughly 5\% of the annotations). 

The annotation task was conducted on Amazon Mechanical Turk (AMT). To ensure the quality of annotations, we required workers to have previously completed 5,000 AMT tasks, and to have an acceptance rate of 98\% or higher. We limited the worker location to the U.S. and Canada, and presented workers with a qualification test similar to the task. We paid 7 cents for each event.

\paragraph{Training.} We fine-tuned GPT-3 on the collected inferences. The input and output format was as follows:\\
\begin{tabular}{l}
     \texttt{Context:} $<$context$>$ \\
     \texttt{Event:} $<$event$>$\\
     \texttt{Before:} $<$before$>$\\
     \texttt{After:} $<$after$>$\\
\end{tabular}

\begin{table*}[t!]
\centering
\setlength{\tabcolsep}{1pt}
\small
\begin{tabular}{P{0.21\textwidth}P{0.41\textwidth}P{0.34\textwidth}}
\toprule
\textbf{Context} & \textbf{Before} & \textbf{After} \\ 
\rowcolor{blue!20} \multicolumn{3}{c}{\textbf{Human-Written}} \\
\multirow{4}{0.21\textwidth}{[...] Chris Weitz will direct the sequel to Twilight, New Moon, \underline{replacing} Catherine Hardwicke.} & 
They couldn't strike a [...] deal with Hardwicke. & Chris Weitz received an advance from the studio. \\
& The executive producer contacted Weitz's agent. & Chris signed the contract. \\
& Weitz's agent communicated the director's message. & His agent put out a press release.  \\
& They decided to replace him. & Chris was happy.\\ \midrule
\multirow{4}{0.21\textwidth}{[...] Chris Weitz will \underline{direct} the sequel to Twilight, New Moon, replacing Catherine Hardwicke.} & The director oversaw the hiring of shooting staff. & People watch the movie. \\
& The director oversaw several screen tests. & He gets paid. \\
& He writes a movie script. & The movie gets released in movie theatres. \\
& He needed to sign a contract. & The movie makes a huge collection. \\
\rowcolor{blue!20} \multicolumn{3}{c}{\textbf{Model-Generated}} \\
\multirow{4}{0.21\textwidth}{Lindsay Lohan \underline{checks} \underline{into} rehab at Betty Ford Center, rehires longtime lawyer Shawn Holley} & 
She decided to change her life from bad to good. & She is treated for her addiction. \\
& She decided to seek help for her addiction. & She attended daily group therapy meetings. \\
& She is assessed by the staff at Betty Ford. & She ends up in the hospital. \\
& She is welcomed by staff members at Betty Ford. & She no longer has a problem. \\ \midrule
\multirow{4}{0.21\textwidth}{Lindsay Lohan checks into rehab at Betty Ford Center, \underline{rehires} longtime lawyer Shawn Holley} & 
Lindsay needs advice on her case. & He gets a good pay. \\
& Lindsay needs legal counsel in her case. & He looks for another case. \\
& she fired her old lawyer. & she went through planning stages of her recovery. \\
& she got a new director. & she started to addiction treatment.\\ 
\bottomrule
\end{tabular}
\caption{Human-written (top) and model-generated (bottom) examples from our multi-event temporal commonsense inference engine. Some examples are slightly abbreviated for readability.}
\label{tab:me_examples}
\end{table*}

Table~\ref{tab:gpt3_examples} shows the format inputted into GPT-3 for training (top row) and inference (bottom row).

\paragraph{Inference} To generate inferences, we prompt the fine-tuned GPT-3 model with the context and the event. We generate up to 150 tokens using top-$p$ decoding \cite{holtzman2020curious} with a cumulative probability of $p=0.9$.  Table~\ref{tab:me_examples} provides examples of the training data (top part) and generated inferences (bottom part) of our multi-event commonsense inference engine. Note, both the human-written and model-generated inferences differ for different events belonging to the same context. For example, according to our model, \textit{after} the event \quotes{Lindsay checks into rehab,} a plausible inference is that \quotes{she gets treated for her addiction.} Yet, \textit{after} the event \quotes{she rehires her longtime lawyer,} our model infers that \quotes{he gets a good pay.} \footnote{We also experimented with prompting GPT-3 in a few-shot setup (Sec \ref{sec:evaluation:ablation}).}
% In the top part of the table, after Chris Weitz is set to replace Catherine Hardwicke as the movie director, his agent puts out a press release, while after Weitz directs the movie, the movie gets released in movie theatres.

% We provide examples of the training data and generated inferences in Appendix~\ref{sec:appendix:gpt3_ft}.

\subsection{Inference-Enhanced Pairwise Scorer}
\label{sec:method:incorporation}
Figure~\ref{fig:pairwise_scorer} shows the overall architecture of our commonsense-enhanced pairwise scorer. We follow \citeauthor{cattan-etal-2021-cross-document}'s mention span representation for mention $m_i$:
\begin{equation}
\label{eq:1}
\operatorname{ctx}_i = [x_{START(i)}, x_{LAST(i)}, \hat{x}_i, l_i]
\end{equation}
\noindent where $x_j$ corresponds to the RoBERTa \cite{roberta-large} embedding of the $j$th token in the span. Each mention is represented as the concatenation of: the first ($x_{START(i)}$) and last ($x_{LAST(i)}$) tokens; an attention-weighted sum of tokens $\hat{x}_i$; and a feature vector denoting the length $l_i$.

%To incorporate the commonsense inference, we generate up to $k=5$ inferences for each of the before ($b$) and after ($a$) relations, using the inference engine: $b_1...b_k$, and $a_1...a_k$.
To incorporate the commonsense inference, we use the inference engine to generate up to $k=5$ inferences for each of the before ($b$) and after ($a$) relations: $b_1...b_k$, and $a_1...a_k$. We describe below the representation of the relation vector using \emph{after} as an example. The representation of the \emph{before} relation is identical.  We first compute the contextualized representation of each inference similarly to the span representations in Equation \ref{eq:1}.

We then stack all the contextualized representations of the inferences:
\begin{equation}
\overrightarrow{A_i} = [\operatorname{ctx}_{a1}...\operatorname{ctx}_{ak}]
\end{equation}
\noindent and input them into a single head attention layer, which produces a single attention-weighted vector for the \emph{after} relation. 

In the context of the pairwise scorer, consider that we have two mention spans $m_i$ and $m_j$ and their corresponding \emph{after} inference representations $A_i$ and $A_j$. We implement two variants of the attention mechanism:
\begin{enumerate}[leftmargin=*,itemindent=.2in,itemsep=.5\parsep]
    \item \emph{Intra-span}, where the attention is between the mention span $m_i$ and the corresponding inferences $\overrightarrow{A_i}$. This is exemplified in Figure~\ref{fig:pairwise_scorer}, where attention is computed between the mention span of \textit{spend} and the inferences corresponding to the same document. The query vector is the mention span $\operatorname{ctx}_i$, and the key vector is the contextualized \emph{after} vector $\overrightarrow{A_i}$. The idea behind this method is to emphasize inferences that are the most relevant to the given mention and provide additional context. 
    \item \emph{Inter-span}, where the attention is between the mention span $m_i$ and the inferences generated for the context of the other mention, $\overrightarrow{A_j}$. For example, in Fig~\ref{fig:pairwise_scorer}, this would mean the purple and orange arrows originating in document 1 would need to be moved to the top row of inferences, corresponding to document 2. The query vector is the span $\operatorname{ctx}_i$, and the key vector is the contextualized \emph{after} vector $\overrightarrow{A_j}$. The goal of this method is to emphasize inferences that are relevant to the other mention, and to bring lexically divergent mentions closer.
\end{enumerate}

In both cases, this leads to an attention-weighted commonsense vector for each of the before and after relations, which are then concatenated to create a single commonsense vector $\operatorname{cs}_i = [\overrightarrow{B_i}, \overrightarrow{A_i}]$ as shown in Figure~\ref{fig:pairwise_scorer}. The input to the pairwise scorer for mentions $m_i$ and $m_j$ is therefore:
\begin{equation}
g_{i,j} = [\operatorname{ctx}_i, \operatorname{ctx}_j, \operatorname{cs}_i, \operatorname{cs}_j]
\end{equation}

The scores from the pairwise scorer are then used to cluster mentions using agglomerative clustering, identically to \newcite{cattan-etal-2021-cross-document}. Agglomerative clustering merges the most similar cluster pairs until their pairwise similarity score falls below a predetermined threshold. 

\section{Experimental Setup}
\label{sec:implementation}
\begin{table*}[t!]
\centering
\resizebox{\textwidth}{!}{
\begin{tabular}{lrrrrrrrrrrr}
\hline
\textbf{Model} & \multicolumn{3}{c}{\textbf{MUC}} & \multicolumn{3}{c}{\textbf{B\textsuperscript{3}}} & \multicolumn{3}{c}{\textbf{CEAFe}} & \multicolumn{1}{c}{\textbf{CONLL}} \\ \hline
\textbf{} &
  \multicolumn{1}{c}{\textbf{P}} &
  \multicolumn{1}{c}{\textbf{R}} &
  \multicolumn{1}{c}{\textbf{F1}} &
  \multicolumn{1}{c}{\textbf{P}} &
  \multicolumn{1}{c}{\textbf{R}} &
  \multicolumn{1}{c}{\textbf{F1}} &
  \multicolumn{1}{c}{\textbf{P}} &
  \multicolumn{1}{c}{\textbf{R}} &
  \multicolumn{1}{c}{\textbf{F1}} &
  \multicolumn{1}{c}{\textbf{F1}} &
  \multicolumn{1}{c}{$\Delta$} \\ \hline
 \textbf{Baseline}   & 73.49     & 84.13     & 78.45    & 48.49     & 67.72     & 56.52     & 43.49   & 55.65        & 48.83     & {61.30 $\pm$ 0.31} & -                    \\ \hline
\textbf{Inter-span} & 74.19     & 84.6      & 79.07    & 50.06     & 68.17     & 57.73    & 44.13   & 55.96          & 49.35     & {62.05 $\pm$ 0.35} & ($\uparrow\text{0.75}$)             \\ \hline
\textbf{Intra-span} & 75.02     & 84.72     & 79.58    & 51.01    & 68.00     & 58.29    & 44.31      & 57.70     & 50.13     & {62.67 $\pm$ 0.24} & ($\uparrow\text{1.37}$)    \\ \hline
% \multicolumn{1}{l}{\textbf{}} &
%   \multicolumn{1}{l}{\textbf{}} &
%   \multicolumn{1}{l}{\textbf{}} &
%   \multicolumn{1}{l}{\textbf{}} &
%   \multicolumn{1}{l}{\textbf{}} &
%   \multicolumn{1}{l}{\textbf{}} &
%   \multicolumn{1}{l}{\textbf{}} &
%   \multicolumn{1}{l}{\textbf{}} &
%   \multicolumn{1}{l}{\textbf{}} &
%   \multicolumn{1}{l}{\textbf{}} &
%   \\
% \multicolumn{1}{l}{\textbf{}} &
%   \multicolumn{1}{l}{\textbf{}} &
%   \multicolumn{1}{l}{\textbf{}} &
%   \multicolumn{1}{l}{\textbf{}} &
%   \multicolumn{1}{l}{\textbf{}} &
%   \multicolumn{1}{l}{\textbf{}} &
%   \multicolumn{1}{l}{\textbf{}} &
%   \multicolumn{1}{l}{\textbf{}} &
%   \multicolumn{1}{l}{\textbf{}} &
%   \multicolumn{1}{l}{\textbf{}} &
%   \textbf{}
\end{tabular}}
\caption{Topic-level performance for event coreference on the ECB+ test set (with gold mentions, no singletons) - Baseline, Inter-span (multi-event commonsense), Intra-span (multi-event commonsense)}
% \caption{Topic Level (Singletons removed): Precision/Recall/$F_1$ scores for Event coreference on ECB+ test set over gold mentions spans}
\label{tab:topic_level_results}
\end{table*}

\subsection{Implementation Details}
\label{sec:experimental:impl}

The implementation of our model is based on \newcite{cattan-etal-2021-cross-document}. We use their official codebase\footnote{\url{https://github.com/ariecattan/coref}} and modify it to support the additional components. Since we use gold event mentions to generate inferences from the multi-event commonsense inference engine (Sec~\ref{sec:method:csmodel}), during both training and inference, we train and evaluate the coreference pipeline on gold mentions. During testing, we evaluate both GPT3 and the coreference system on new gold mentions that are not seen during training. This is in contrast to \newcite{cattan-etal-2021-cross-document} which learned to extract candidate mention spans and train the coreference system. However, using gold mentions is common practice among many coreference systems where the focus is on improving the pairwise scorer \cite[e.g.][]{barhom-etal-2019-revisiting,yadav-etal-2021-event}. For a fair comparison, we report the baseline performance by re-running \citet{cattan-etal-2021-cross-document} using gold mentions similar to the baseline used in \citet{yadav-etal-2021-enhancing}.
We compare this baseline to two variants of our model, based on intra-span and inter-span attention (Sec~\ref{sec:method:incorporation}). We train all model versions using 15 different random seeds, and we report the average performance. 

For our GPT-3 based inference engine, we fine-tuned the \textit{Davinci} model which we accessed via the OpenAI API.\footnote{\url{https://beta.openai.com/}}. The hyperparameters for all the models are detailed in Appendix~\ref{sec:appendix:hyperparams}.

\subsection{Evaluation Setup and Metrics}
\label{sec:experimental:eval}
% he evaluation method 
% \cite{cattan-etal-2021-cross-document} pruned document spans down to the gold mentions 
The primary metric we use is the standard CONLL-$F_1$ implemented by \newcite{moosavi-strube-2016-coreference}\footnote{\url{https://github.com/ns-moosavi/coval}}, which is the average of three metrics: B\textsuperscript{3} \cite{Bcubed}, MUC \cite{vilain-etal-1995-model}, and CEAF\textsubscript{e} \cite{luo-2005-coreference}. We follow the evaluation setup used in recent work \cite{cattan-etal-2021-cross-document, yadav-etal-2021-event, held-etal-2021-focus, Cattan2021RealisticEP} and evaluate all our models at the topic level. That is, each metric is computed for each topic separately and averaged across all topics. We also remove singleton clusters (clusters with a single mention) as they have shown to artificially boost the scores when using gold mentions \cite{cattan-etal-2021-cross-document}. 
% While the topic-level setup has faced some criticism \cite{upadhyay-etal-2016-revisiting}, recent work \cite{Cattan2021RealisticEP}, as it represents a more realistic scenario in which documents are already clustered according to topics and only mention pairs within the same topic are considered potentially coreferring \cite{cattan-etal-2021-cross-document}. It is important to note that each topic is composed of sub-topics with different seminal events, making the topic-level setup challenging as well.
% \subsection{Baselines}
% \label{sec:experimental:baselines}
% Our baseline is the method proposed by \newcite{cattan-etal-2021-cross-document}. For fair comparison, we re-ran the system using gold mentions. We compare this baseline to two variants of our model, based on interspan and interspan attention. All our models are trained using 10 different random seeds, and report the average and standard deviation for each metric. The hyperparameters for all our models are based on \newcite{cattan-etal-2021-cross-document}, and are detailed in Appendix~\ref{sec:appendix:hyperparams}. 

\section{Evaluation}
\label{sec:eval}
We discuss the results on the event coreference task (Sec~\ref{sec:evaluation:results}), the validity of the commonsense inferences generated by our inference engine (Sec~\ref{sec:evaluation:human}), and present ablation tests (Sec~\ref{sec:evaluation:ablation}).
\subsection{Results}
\label{sec:evaluation:results}
Table~\ref{tab:topic_level_results} shows the performance of the baseline and the inter-span and intra-span variants of the proposed multi-event commonsense models on event coreference on the ECB+ test set. 
Both of our proposed variants improve upon the baseline in terms of CONLL-$F_1$, with our intra-span model yielding an increase of 1.37 ($\pm$ 0.24) points, and our inter-spam model yielding an increase of 0.75 ($\pm$ 0.35).
%Our proposed variants improve upon the baseline in terms of CONLL-$F_1$, with intra-span performing better, with an increase of 1.37 ($\pm$ 0.24) points from the baseline, compared to an increase of 0.75 ($\pm$ 0.35) for the inter-span model.
Overall, the improvement in performance indicates that the temporal commonsense inferences helped in resolving a considerable number of coreferences, which we analyze in more detail in Sec~\ref{sec:analysis:attention}. In particular, both models improve upon the baseline precision across all metrics, with the intra-span model achieving the highest precision across all metrics. Error analysis of the best model (intra-span, Sec~\ref{sec:analysis:errors}) shows that in some cases when mentions had similar (and possibly generic) inferences, the model falsely classified non-coreferring mentions as coreferring.
%Error analysis of the best model (intra-span, Sec~\ref{sec:analysis:errors}) shows that in some cases, the model falsely classified non-coreferring mentions as coreferring, when they had similar (possibly generic) inferences.
We hypothesize that this error is more common for the inter-span model. When one mention's inference is lexically similar to the other mention, it would get more attention, increasing the likelihood of a false positive error.

\subsection{Human Evaluation of Inferences}
\label{sec:evaluation:human}
We manually evaluate the quality of the commonsense inferences generated by our inference engine (Sec~\ref{sec:method:csmodel}). We randomly sampled 600 inferences from the validation set. We used the same AMT qualifications as in Sec~\ref{sec:method:csmodel} and paid 20 cents per HIT.\footnote{See Appendix~\ref{appendix:he} for the HIT template.} We presented three workers with a sentence and a target event, followed by the before and after inferences generated by the model. We asked them about the inference's (i) \emph{likelihood}, i.e. how often would the given inference actually occur before (after) the target event; (ii) \emph{relevance} with respect to the context; and (iii) \emph{specificity} of the inference with respect to the target event. Table~\ref{tab:humaneval} presents the results. As expected, the generated inferences were almost always relevant to the corresponding event contexts. The majority of inferences (78.8\%) were specific to the target event, but there was a significant percent of moderately specific inferences (19.4\%) that often pertained to other events in the context. Finally, the majority of inferences either always (58\%) or sometimes (36.1\%) happen before or after the target event. These results reconfirm the extrinsic gains in Sec~\ref{sec:evaluation:results}, and suggest that the inference engine may be useful for other NLP tasks. The inter-annotator agreement in terms of Fleiss kappa for the three metrics are as follows: Likelihood = 0.71, Relevance - 0.65, and Specificity - 0.84 (substantial agreement).
\subsection{Ablation Tests}
\label{sec:evaluation:ablation}
In Sec~\ref{sec:method:csmodel}, we argued that COMET is insufficiently accurate for complex sentences with multiple events. To collect evidence, we replace our GPT-3 based commonsense inference engine with COMET and re-train the event coreference model. We used the newest COMET version \cite{Hwang2021COMETATOMIC2O}, along with beam search to decode the top 5 inferences for each relation type (before/after), ranked based on the model's confidence. 

In addition, to justify fine-tuning GPT-3, we also replace our multi-event commonsense inference engine with a few-shot version of the model. We randomly sampled 8 of the human-written inferences (Sec~\ref{sec:method:csmodel}) to prompt GPT-3, and we used the same instructions to prompt it to generate before and after inferences. In all experiments, the rest of the model is as described in Sec~\ref{sec:method:incorporation}. 

Table~\ref{tab:event-knowledge} presents the ablation results. The COMET-based model shows a marginal improvement from the baseline, yet performs worse than the multi-event inference engine. The few-shot GPT-3 model performs better, but we discovered that more training data could improve the specificity and accuracy of the inferences. Finally, our fine-tuned GPT-3 inference engine outperforms all models, thanks to its explicit training on multi-event inferences.
\begin{table}[t!]
\centering
\resizebox{\columnwidth}{!}{%
\begin{tabular}{llrrr}
\hline
\textbf{} & \textbf{Metric} & \multicolumn{1}{l}{\% High} & \multicolumn{1}{l}{\% Moderate} & \multicolumn{1}{l}{\% Low} \\ \hline
1. & Likelihood        & 58.0 & 36.1 & 6.1 \\
2. & Relevance & 97.3 & 0.0  & 2.7 \\
3. & Specificity       & 78.8 & 19.4 & 1.8 \\
\hline
\end{tabular}%
}
\caption{Human evaluation results for the inferences generated by our commonsense inference engine.}
\label{tab:humaneval}
\end{table}
\begin{table}[t]
\centering
\small 
\begin{tabular}{lll}
\toprule
\textbf{Model} & \textbf{Inter-span} & \textbf{Intra-span} \\ \midrule
Baseline (no inf.) & \multicolumn{2}{c}{61.3 $\pm$ 0.31} \\
COMET & {61.51 $\pm$ 0.21} & {61.39 $\pm$ 0.32} \\
GPT-3 few-shot & {61.59 $\pm$ 0.26}  & {61.64 $\pm$ 0.35}\\
GPT-3 FT (ours)  & {62.05 $\pm$ 0.35} & {62.67 $\pm$ 0.24} \\ 
\bottomrule
\end{tabular}
\caption{CONLL-F$_1$ performance on the ECB+ test set using different event commonsense knowledge sources.}
\label{tab:event-knowledge}
\end{table}

\section{Analysis}
\label{sec:analysis}

\begin{figure*}[t!]
    \centering
    \includegraphics[width=.7\textwidth]{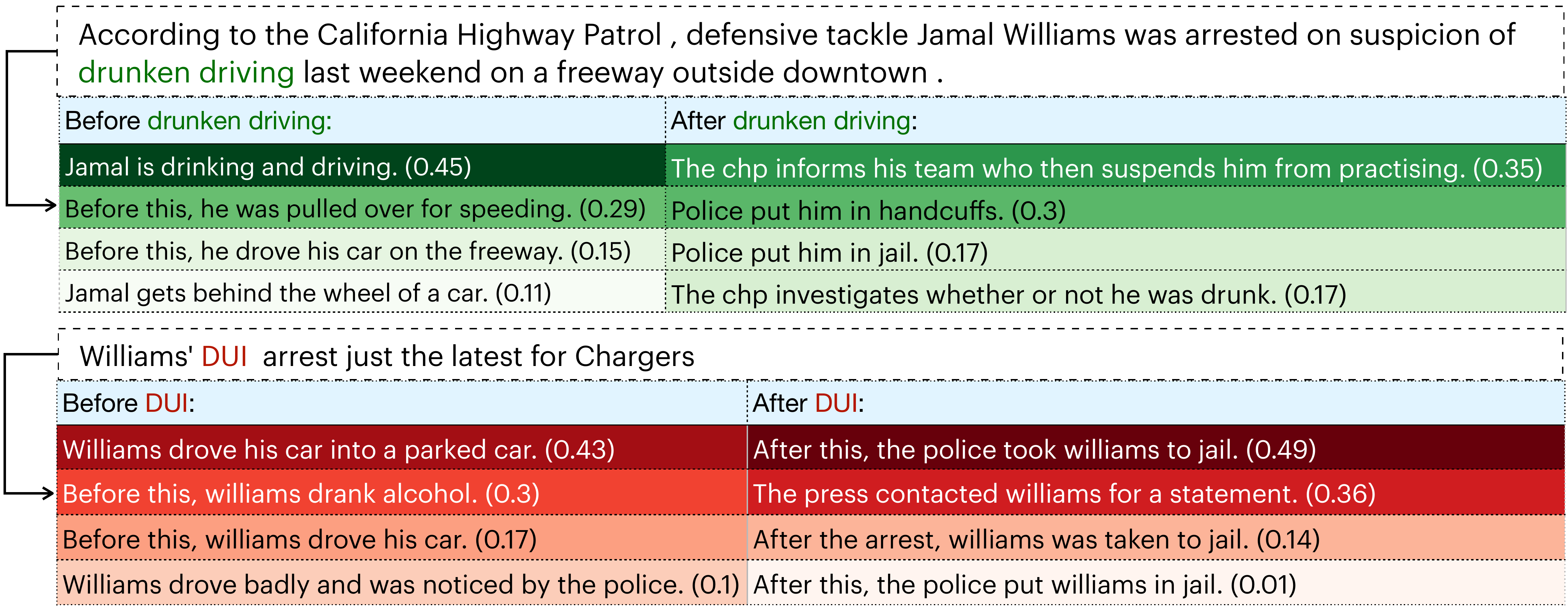}
    \caption{An example mention pair and the intra-span attention weights between the contexts and the inferences.}
    \label{fig:attention_intraspan}
\end{figure*}
\begin{figure*}[t!]
    \centering
    \includegraphics[width=.7\textwidth]{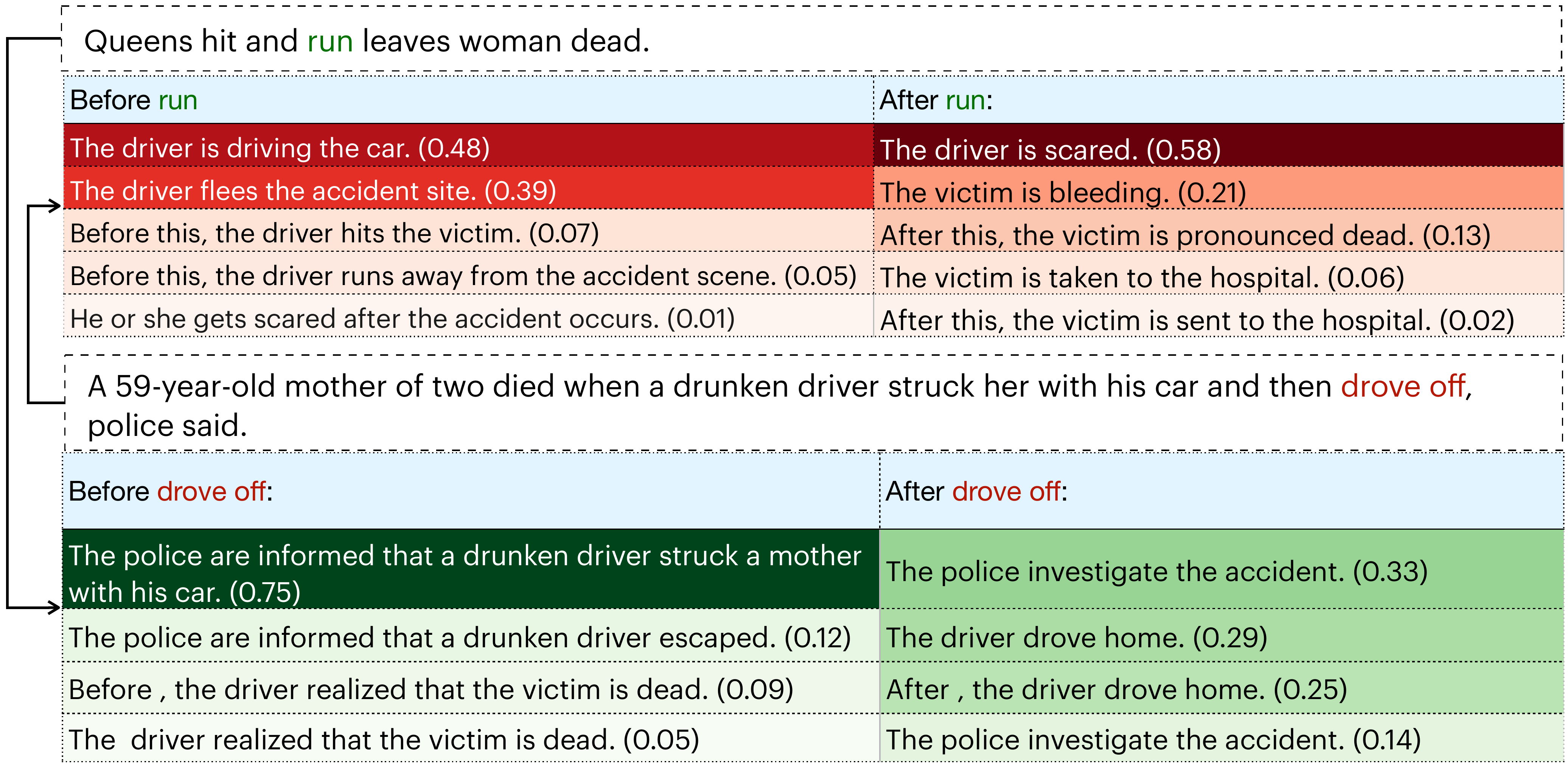}
    \caption{An example mention pair and the inter-span attention weights between the contexts and the inferences.}
    \label{fig:attention_interspan}
\end{figure*}

\subsection{Attention Scores}
\label{sec:analysis:attention}
Figure~\ref{fig:attention_intraspan} presents an example of a mention pair (\textit{drunken driving}, \textit{DUI}) that was incorrectly predicted as non-coreferring by the baseline and correctly predicted as coreferring by the intra-span model. The inferences for each mention are sorted and highlighted according to their corresponding attention weights. The highest scoring \emph{before} inference for the first mention, ``Jamal is drinking and driving'', and the second inference for the second mention ``Wiliams drank alcohol'', are similar, which likely contributed to recognizing the coreference. 
Figure~\ref{fig:attention_interspan} similarly shows an example that was incorrectly predicted as non-coreferring by the baseline and correctly predicted as coreferring by the inter-span model. Here, we can clearly observe the interplay between the second mention \textit{drove off} and the inferences of the first mention \textit{hit} related to driving and fleeing from the scene. The lexical and contextual diversity of these mentions necessitates commonsense inferences to resolve the coreference. Appendix~\ref{sec:appendix:attention_weights} provides a second set of examples.

\begin{table}[t]
\centering
\small
\begin{tabular}{lr}
\toprule
\textbf{Category}  & \textbf{\%} \\ \midrule
\textcircled{1} Lack of Structure &	29.5\\
\textcircled{2} Generic Inferences &	24.6\\
\textcircled{3} Insufficient Knowledge & 19.5\\
\textcircled{4}  Incorporation &	18.1\\
\textcircled{5}  Attention &	8.3\\
\bottomrule
\end{tabular}
\caption{Error analysis of the intra-span model.}
\label{tab:error_analysis}
\end{table}
% \vspace{-7pt}

\subsection{Error analysis}
\label{sec:analysis:errors}

We analyze the errors in the best version of our model (intra-span). 95\% of the errors made by this model overlapped with the errors made by the baseline, and only 5\% were newly-introduced. We sampled 100 errors from the validation set and manually categorized them into the following categories, detailed below and quantified in Table~\ref{tab:error_analysis}. See Appendix~\ref{appendix:error_examples} for examples from each category.
\begin{enumerate}[label=\large\protect\textcircled{\small\arabic*},leftmargin=*,itemindent=.2in,itemsep=.5\parsep]

\item \textbf{Lack of Structure}: Similar or identical mentions may refer to different events, as in \quotes{\emph{Jackman} hosting the Academy awards} vs. \quotes{\emph{Ellen} hosting the Oscars}. Previous work incorporated semantic roles into the mention representation to identify such cases  \cite{barhom-etal-2019-revisiting}. Our baseline model, as well as the inferences from our proposed approach, do not explicitly incorporate any linguistic structure, which results in these errors.

\item \textbf{Generic Inferences}: The generated commonsense inferences are not specific enough with respect to the target event. This causes both false positive errors, when a pair of non-corefering mentions have similar generic inferences; and false negative errors, when coreferring mentions have dissimilar generic inferences. 

% \item \textbf{Generic Inferences}: 
% These errors occur because, the inferences confuse and mix up different events within the same context. Inferences that are not specific enough can lead to corefering mention pairs having less overlap (\eg Inferences of \quotes{shot} in Figure~\ref{fig:architecture} may be specific while inferences of \quotes{gunshot} may be mixed up with \quotes{hospitalized}) or non-corefering mentions having higher overlap (false positive).
% The inferences were generic rather than event-specific, so the two non-coreferring mentions had similar generic inferences.   
  
\item \textbf{Insufficient Knowledge}: The inferences are relevant to the target event, but don't contain all the knowledge required to resolve these coreferences.  % (\eg They require more relations than \emph{before} or \emph{after})
  
\item \textbf{Incorporation}: The inferences and attention scores were accurate, but the model did not use them effectively during incorporation. %These may be due to the model not being able to understand the relevance of the inferences.
% \vered{I rephrased but I'm not sure about the description, it's a bit confusing.}

\item \textbf{Attention}: The model either attended too much to unnecessary inferences (weights close to 1) or ignored crucial inferences (weights close to 0).

% mentions have lexical or other similarities in inferences despite being non-coreferring.  It also includes cases where inferences are relevant, but have lexical dissimilarity in the inferences.
\end{enumerate}
% \label{appendix:error_examples}

\section{Conclusions}
\label{sec:conclusions}
In this paper, we investigated the effect of injecting temporal commonsense knowledge in the task of event coreference resolution. By using event-specific inferences generated by our commonsense model, we improve the performance of a baseline model. Our analysis shows that the pairwise scorer attends to inferences that are beneficial in solving challenging coreferences. In the future, we plan to extend the multi-event commonsense model to additional relations, and to incorporate such knowledge into other discourse tasks.

%As shown in our analysis, the attention-based incorporation of inferences proved beneficial in solving challenging coreferences. Our analysis also demonstrated the challenges involved in incorporating reasoning in discourse tasks. In the future, we plan to work towards adding more relations to the multi-event commonsense model and exploring improved incorporation methods for discourse tasks.

\section{Limitations}
\label{sec:limitations}
\paragraph{Data.} As shown by \newcite{barhom-etal-2019-revisiting}, ECB+ suffers from annotation errors. In particular, the event coreference annotations are incomplete, which might lead to false positive errors for truly coreferring mention pairs. In this work, we intentionally addressed the edge cases in event coreference that haven't been addressed by prior research: lexically/contextually-divergent mentions. The number of such corefering clusters in ECB+ is small, and it has been shown that just clustering together mention pairs with the same lemma yields an F1 score of 42.3 on the dataset \cite{upadhyay-etal-2016-revisiting}. Further, our analysis of corefering pairs on the validation set revealed that only 11\% of the pairs were contextually dissimilar (cosine similarity  below 0.9), indicating that commonsense may impact only these cases. Unfortunately, this is the standard dataset for event coreference, but in the future, we could think of collecting a more challenging (and realistic) dataset. 
\paragraph{Models.} The accuracy of the commonsense model is primarily limited by the accuracy of inferences from GPT-3. \newcite{marcus2020experiments} tested GPT-3 on various types of commonsense reasoning and found mixed results for temporal commonsense. Our human evaluation in Sec~\ref{sec:evaluation:human} revealed that GPT-3 generates inferences that are not specific enough to the target event in 19.3\% of the cases, which decreases performance as shown in Sec~\ref{sec:analysis:errors}. We aim to address this in future work by building a more robust multi-event commonsense engine. Another error our model doesn't address concerns semantic roles, which happens when the main difference is in the person, time or location (e.g. two earthquake reports in different times and locations) \cite{barhom-etal-2019-revisiting}.
\paragraph{Evaluation.} Since our commonsense engine was trained with gold event mentions, we used gold mentions to evaluate the coreference model as well. Using predicted mentions instead of gold mentions would provide a more realistic estimate of the performance of an event coreference system. With that said, our work focused on improving the coreference decisions; hence, we followed previous work and used the gold mentions \cite{barhom-etal-2019-revisiting,held-etal-2021-focus}. 

% Finally, following \newcite{cattan-etal-2021-cross-document}, we report the performance at the topic level rather than corpus level, which may be a more optimistic estimate as the topics are already provided.

\section{Acknowledgements}
\label{ack}
This work was funded, in part, by the Vector Institute for AI, Canada CIFAR AI Chairs program, an NSERC discovery grant, and a research gift from AI2.

\bibliography{anthology,custom}
\bibliographystyle{acl_natbib}

\section{Appendix}
\label{sec:appendix}
\appendix

\section{Multi-Event Commonsense Inference Engine}
\label{sec:appendix:gpt3_ft}
\begin{table*}[t!]
\centering
\small
\ttfamily
\begin{tabular}{P{\textwidth}}
\toprule
Context: Rumored to be the front runner earlier in the week , Entertainment Weekly has now confirmed that Chris Weitz will direct the sequel to Twilight , New Moon , replacing Catherine Hardwicke. \\ ~\\
Event: replacing \\
~\\
Before: They could not strike a favorable deal with Catherine Hardwicke. They decided to replace him. Before this, the film's executive producer contacted Weitz's agent. Before this, Weitz's agent communicated the director's message.\\
~\\
After: Chris Weitz received an advance from the studio. Chris signed the contract. After, his agent put out a press release. Chris was happy. END\\ \midrule
Context: Lindsay Lohan checks into rehab at Betty Ford Center , rehires longtime lawyer Shawn Holley \\
~\\
Event: rehires \\
\bottomrule
\end{tabular}
\caption{Examples of the input format of the multi-event commonsense inference engine. Top: a training example is fed into GPT-3 with the inputs (context and event) and the outputs (before and after inferences). Bottom: a test example is fed with only the inputs (context and event).}
\label{tab:gpt3_examples}
\end{table*}

\section{Hyper-Parameters}
\label{sec:appendix:hyperparams}
\begin{table}[ht]
\centering
\small
\begin{tabular}{lr}
\toprule
\textbf{Parameter}  & \textbf{Value} \\ \midrule
Batch Size & 128 \\
Learning Rate & 0.0001 \\ 
Dropout  & 0.3 \\ 
Optimizer & Adam\\ 
Hidden layer & 1024  \\ 
Attention heads & 1 \\
\bottomrule
\end{tabular}
\caption{Hyperparameters used by all three model versions-Baseline, Inter-span and Intra-Span}
\label{tab:hyperparameters}
\end{table}
Table~\ref{tab:hyperparameters} shows the hyperparameters used by all our models. It took an average time of \emph{120} minutes to run the entire pipeline for the inter-span and inter-span versions, and an average time of \emph{80} minutes for the baseline version. We used a single NVIDIA GeForce GTX 1080 Ti GPU for each run. It took $5$ minutes to fine-tune the GPT-3 Davinci model and costed 170 USD for training and generating all the inferences.

\section{Attention Scores}
\label{sec:appendix:attention_weights}

\begin{figure*}
    \centering
    \includegraphics[width=.8\textwidth]{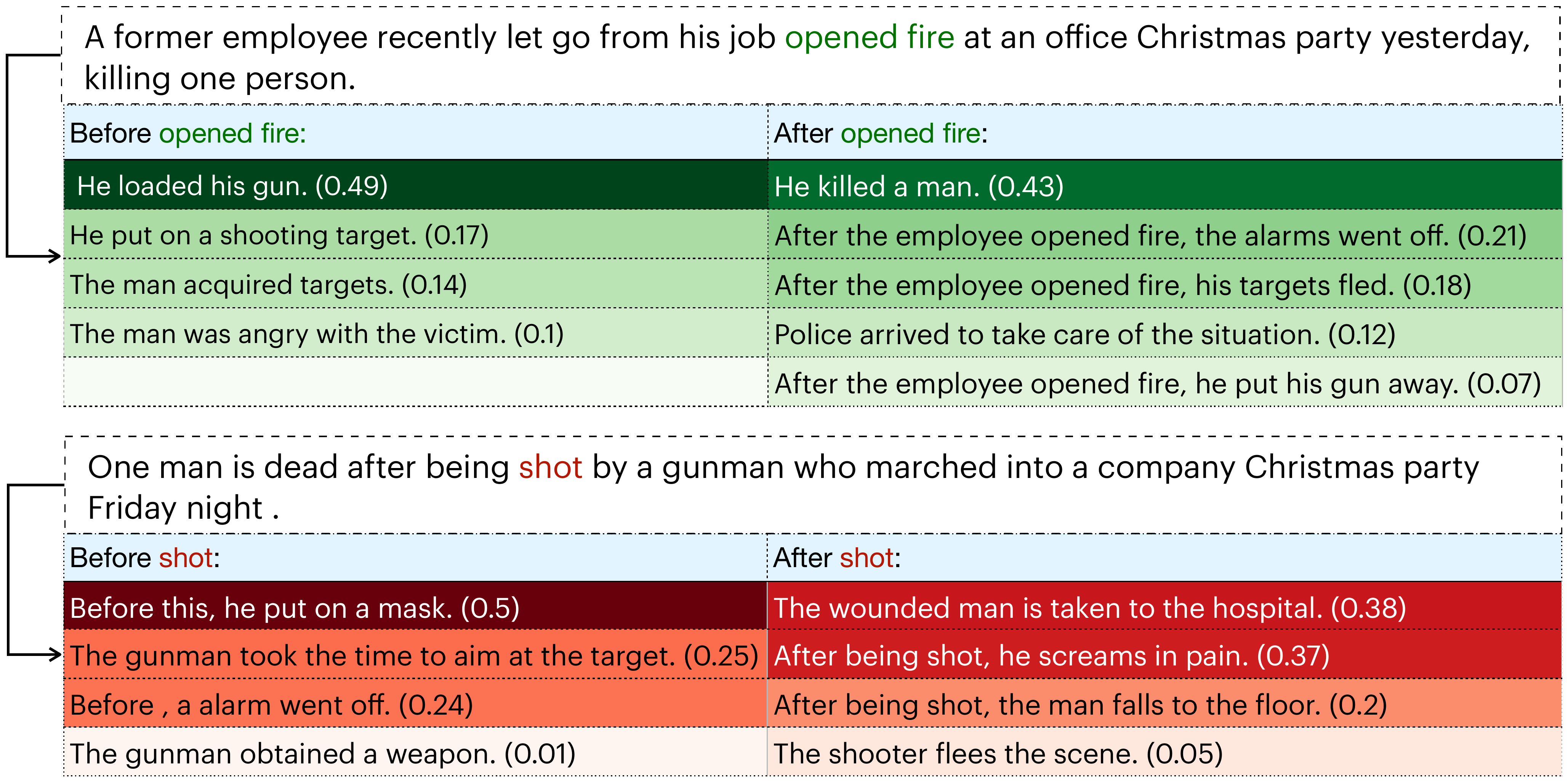}
    \caption{An example mention pair and the intra-span attention weights between the contexts and the inferences.}
    \label{fig:attention_intraspan_appendix}
\end{figure*}

\begin{figure*}[t]
    \centering
    \includegraphics[width=.8\textwidth]{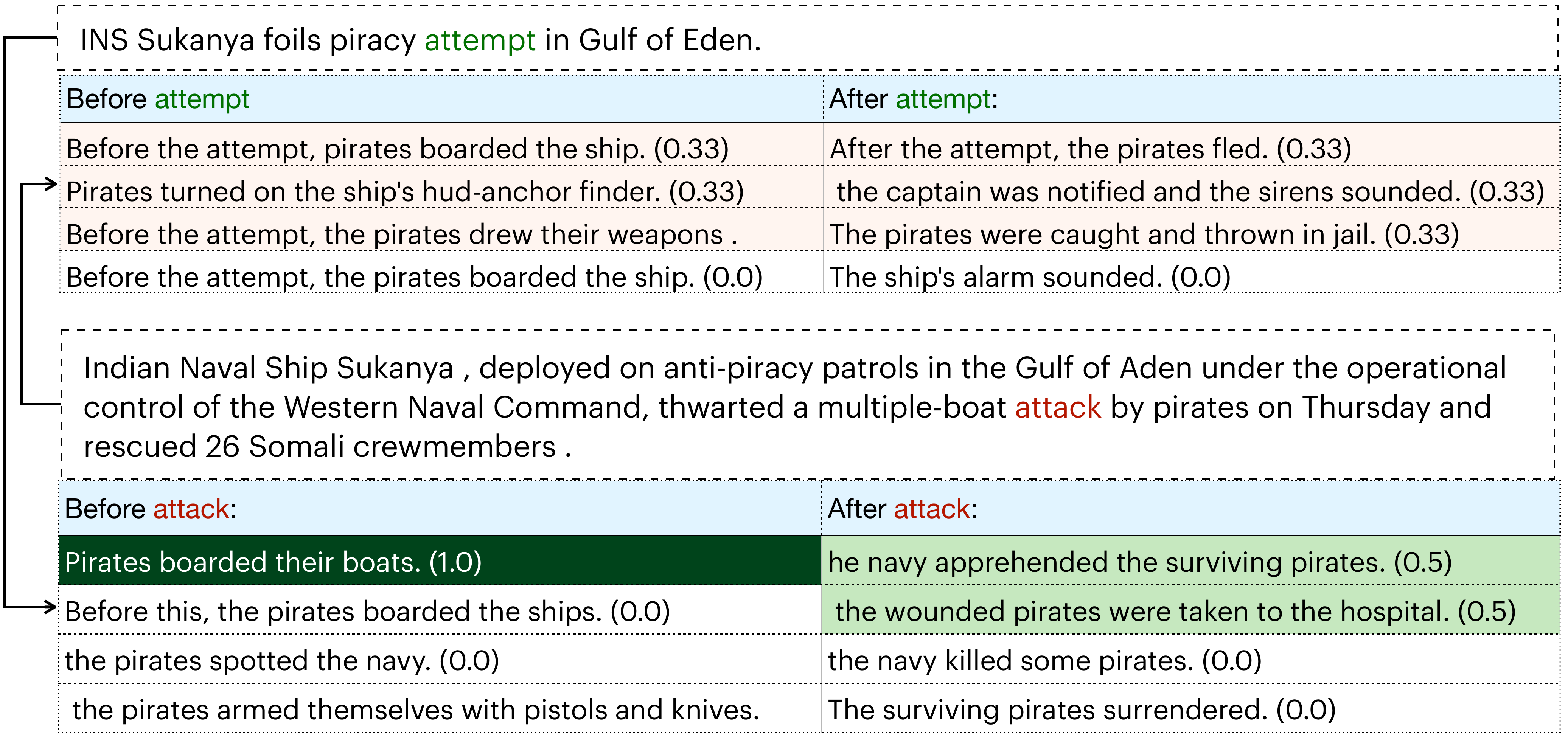}
    \caption{An example mention pair and the inter-span attention weights between the contexts and the inferences.}
    \label{fig:attention_interspan_appendix}
\end{figure*}

In Figures \ref{fig:attention_intraspan_appendix} and \ref{fig:attention_interspan_appendix}, we provide examples for mention pairs incorrectly predicted by the baseline and correctly predicted by the intra-span and inter-span models, repsectively, similarly to Sec~\ref{sec:analysis:errors}.

\section{Error Analysis}
\label{appendix:error_examples}
\begin{table*}[t!]
\centering
\setlength{\tabcolsep}{1pt}
\footnotesize
\begin{tabular}{P{0.21\textwidth}P{0.37\textwidth}P{0.37\textwidth}}
\toprule
\textbf{Context} & \textbf{Before} & \textbf{After} \\ 
\rowcolor{blue!20} \multicolumn{3}{c}{\textcircled{1} \textbf{Lack of Structure: Different arguments (Robert Buckley vs Duncan Rait), similar event mentions and contexts.}} \\
 %% First
\multirow{4}{0.21\textwidth}{Robert Buckley the second climber to \underline{die} in the Aoraki - Mount Cook national park.} &
He slipped and fell down. (0.68) & The bodies of the climbers are found by the other climbers. (1.0)\\
&He set out to climb a mountain. (0.19) & The families of the two climbers are notified. (0.0)\\
&He was exhausted from the trek. (0.13) & the police investigate his cause of death. (0.0)\\
 \midrule
 %% Second
\multirow{4}{0.21\textwidth}{The day before Buckley's death another climber Duncan Rait, \underline{died} after slipping and falling [...].} 
&Before this, the climber slipped and fell. (0.5) & A rescue team went to look for the climber. (0.5)\\
&The climber had an accident. (0.5)& A funeral is held for the dead body. (0.5)\\
&Before this, the climber sustained injuries. (0.0)& the climber was pronounced dead. (0.0)\\
& & \\
%%%%%%%%%%%%%%%%%%%%%%%
\rowcolor{blue!20} \multicolumn{3}{c}{\textcircled{2} \textbf{Generic error - Inferences of the first event (crash) are not specific and accurate.}} \\
 %% First
\multirow{4}{0.21\textwidth}{Man charged with DWI , leaving scene after S . Rich Hill mother killed in \underline{crash} : NYPD} &
They spotted a car on fire. (0.35) & They called the fire department. (0.41)\\ &
She swerved to avoid a cat crossing the road. (0.33) & Her family had to deal with the death. (0.34) \\ &
The driver got into an accident. (0.22) & Police arrived on the scene. (0.18) \\ &
The car collided with the rich hill mother. (0.05)& She was taken to the hospital. (0.07)\\ &
The rich hill mother is driving her car. (0.05) & \\
 \midrule
  %% Second
\multirow{4}{0.21\textwidth}{Woman Killed in Queens Hit - \underline{Run} , Driver Charged } & 
He changed his mind and decided to go forward with the plan. (0.46) & The woman is killed. (0.64) \\ &
A driver wants to kill the woman. (0.27) & They file a case against the driver. (0.11) \\ &
A driver sees the woman. (0.15) & The driver gets worried about the consequence. (0.1) \\ &
He gets scared and attempts to flee the scene. (0.06) & The woman is denied basic rights. (0.08) \\ &
He flees the scene. (0.06) & The woman is denied a burial. (0.08) \\ 
%%%%%%%%%%%%%%%%%%%%%%%
%%%%%%%%%%%%%%%%%%%%
\rowcolor{blue!20} \multicolumn{3}{c}{\textcircled{3} \textbf{Insufficient knowledge error- More knowledge may be beneficial (\eg pre-requisites of events)}} \\
%  %% First
\multirow{4}{0.21\textwidth}{MSNBC is reporting that the Indian Navy claims they have \underline{captured} 23 pirates in the Gulf of Aden} &
The navy ships noticed the pirates. (0.33) & The captured pirates were taken to prison. (0.25) \\&
They boarded the ship. (0.33) & They will decide what to do with them. (0.25) \\&
The captain ordered an alert. (0.33) & the captain signaled the all-clear. (0.25)\\&
The navy ships surrounded the pirates. (0.0) &  The navy notified the police about the capture. (0.25)\\
 \midrule
  %% Second
\multirow{4}{0.21\textwidth}{The Indian Navy on Saturday prevented pirates from attacking a merchant vessel[..] \underline{took} 23 into custody. 
  } & 

They planned to attack the ship. (0.33) & The navy handed them over to the police. (0.33)\\&
The pirates hid their weapons. (0.33) & The navy interrogate them (0.33)\\&
The navy received a distress call from the ship. (0.33) & The navy took them to a different place (0.33)\\

%%%%%%%%%%%%
\rowcolor{blue!20} \multicolumn{3}{c}{\textcircled{4} \textbf{Incorporation error - Inferences seem relevant, but the model fails to use them.}} \\
 %% First
\multirow{4}{0.21\textwidth}{5 Thoughts on Why the Academy Picked Ellen DeGeneres As \underline{Oscar} Host } &
Ellen accepted to host the Oscars. (0.36) & Ellen feels happy(0.34)\\ 
& Ellen was practicing out ideas.(0.36) & The host gets paid. (0.26)\\
 & Ellen DeGeneres was selected as the host. (0.19) & Ellen was given a plaque of honor. (0.22)\\ 
& They academy contacted Ellen(0.09)\\& The audience clapped for Ellen. (0.18)\\
 \midrule
  %% Second
\multirow{4}{0.21\textwidth}{It will be her second stint in the job , after hosting the 2007 ceremony and earning an Emmy \underline{nomination} for it} & 
She practiced her speech. (0.32) & The press contacted her for interviews. (0.55)\\ &
She contacted her suppliers about a new gown for the show. (0.48) & She was very happy. (0.23)\\ &
She was effective in her duties. (0.1) & She informed her staff about the nomination. (0.12) \\ &
She was nominated for hosting the 2007 ceremony. (0.05) & She bought some new clothes. (0.1)\\
 %% %%%%%%%%%%
 
\rowcolor{blue!20} \multicolumn{3}{c}{\textcircled{5} \textbf{Attention error: Increased attention on irrelevant inferences (first inference)}} \\
 %% First
\multirow{4}{0.21\textwidth}{Woman Killed in Queens \underline{Hit} - Run , Driver Charged} &
The driver came into contact with the woman. (1.0) & The driver flees the scene of the collision. (1.0)\\
&The person driving a vehicle saw the woman and  pursued, not caring about the person's safety. (0.0) & The woman is injured. (0.0)\\
&The driver and the woman crossed paths.(0.0) & The woman is hospitalized. (0.0)\\
&The driver drove his vehicle at the woman. (0.0) & The driver tried to hide his involvement in the crime. (0.0)\\
 \midrule
 %% Second
\multirow{4}{0.21\textwidth}{Cops : Queens Woman Killed In Hit - And - \underline{Run}} & 
A car flees the scene. (0.34) & They put out an alert to look for him. (0.28)\\
& A car crashes into a dying woman. (0.26) & They put out a press release calling for information. (0.28)\\ &
They searched the area the car was spotted in. (0.21) & They arrested him. (0.22)\\
& They interviewed neighbors who might have seen them. (0.2) & The criminal went to court  (0.22)\\
%%%%%%%%%%%%%%%%%%%%%%%

\bottomrule
\end{tabular}
\caption{An example of each error category described in Sec~\ref{sec:analysis:errors}}
\label{tab:error_categories}
\end{table*}

Table~\ref{tab:error_categories} shows one example of each error category described in Sec~\ref{sec:analysis:errors}.

% \section{COMET}
% \label{appendix:comet}
% \sahi{add parameters used to obtain inferences from COMET.}

\section{Crowdsourcing Templates}
\label{appendix:he}
\begin{figure*}[ht]%
\centering
\includegraphics[width=0.8\textwidth]{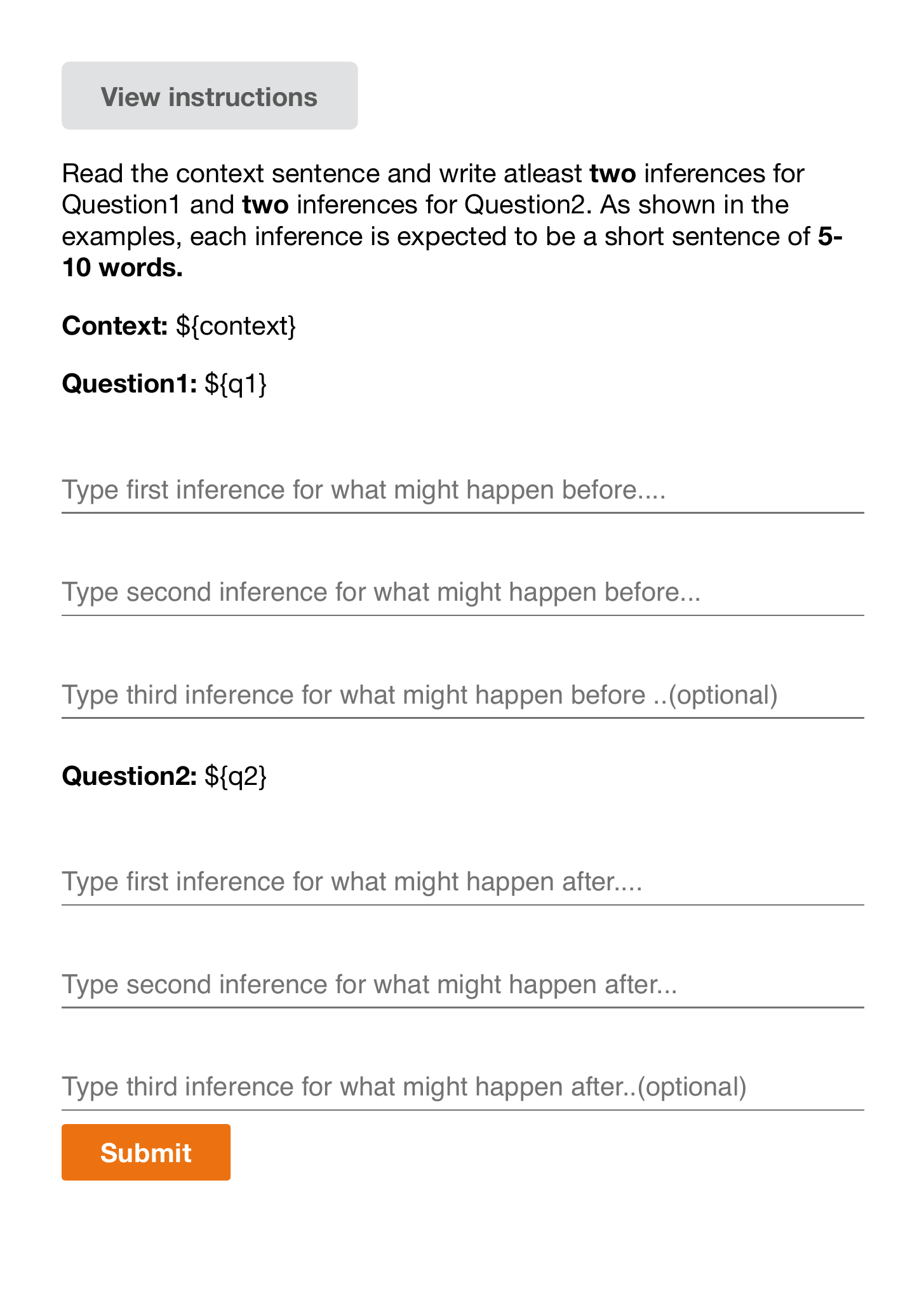}%
\caption{Crowdsourcing template for obtaining before and after inferences.}%
\label{fig:layout1}%
\end{figure*}
\begin{figure*}[ht]%
\centering
\includegraphics[width=\textwidth]{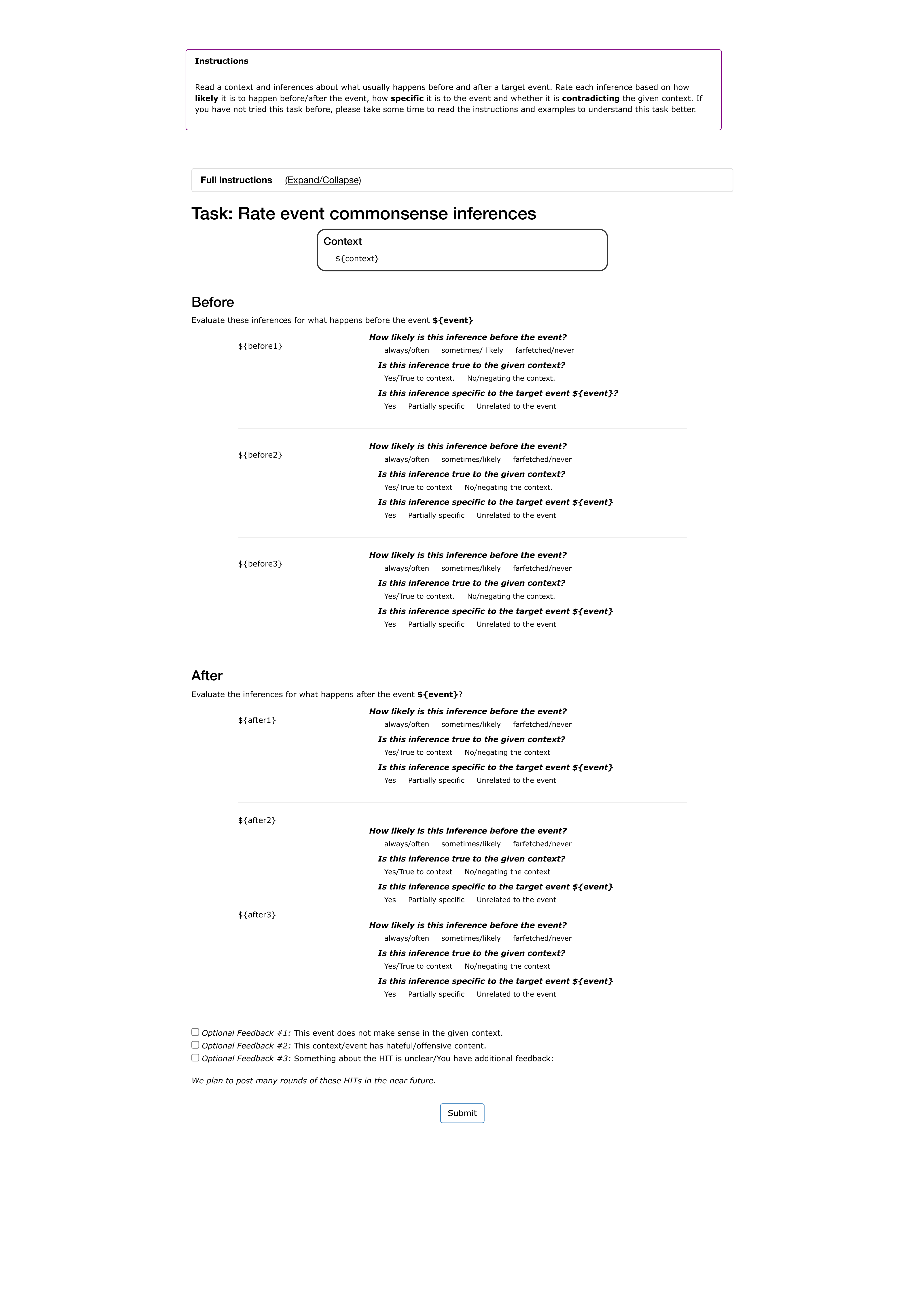}%
\caption{Crowdsourcing template for rating before and after inferences.}%
\label{fig:layout2}%
\end{figure*}
Figures \ref{fig:layout1} and \ref{fig:layout2} show the HIT templates used for obtaining inference annotations and evaluating generated inferences, respectively.

% Our annotation as well as evaluation is conducted on Amazon Mechanical Turk. To ensure the quality of annotations, we required that workers have completed prior tasks with a minimum acceptance rate of 90\%. We limited the worker location to US and Canada, and presented workers with a qualification test similar to the task. We paid $7$ cents for each example for annotations and $0.2$ cents per example for human evaluation. \sahi{Add figure to show human evaluation template.}

\end{document}